\documentclass[runningheads]{llncs}
\usepackage[T1]{fontenc}
\usepackage[utf8]{inputenc}
\usepackage[table]{xcolor}
\definecolor{darkpastelgreen}{rgb}{0.01, 0.75, 0.24}
\usepackage{graphicx,verbatim}
\usepackage{booktabs}
\usepackage{float} 
\usepackage[breaklinks,colorlinks,citecolor=black, linkcolor=black, urlcolor=blue]{hyperref}
\usepackage{cleveref}
\usepackage[table]{xcolor}
\usepackage{multirow}
\usepackage{subcaption}
\usepackage{ulem}
\usepackage{amssymb}

\usepackage{xcolor}
\newcommand{\bstar}{\textcolor{blue}{\ensuremath{^{\star}}}}   
\newcommand{\ystar}{\textcolor{orange!90!black}{\ensuremath{^{\star}}}} 

\begin{document}
\title{Prompt-Guided Interactive Segmentation of Interstitial Lung Disease in Thoracic CT}
\titlerunning{Prompt-Guided Interactive Segmentation of ILDs in CT}
%
\author{
Vasilis Dedousis\inst{1,2}$^{\dagger}$ \and
Lubnaa Abdur Rahman\inst{1,2}$^{\dagger}$ \and
Lorenzo Brigato\inst{1} \and
Ethan Dack\inst{1} \and
Andreas Christe\inst{3} \and
Christoph Frank \inst{4} \and
Manuela Funke-Chambour\inst{3,5} \and
Justus Roos \inst{4}  \and
Adrian Huber \inst{4} \and
Lukas Ebner\inst{3,4} \and
Stavroula Mougiakakou\inst{1,3}
}

\authorrunning{Dedousis, Abdur Rahman, et al.}
%

\institute{University of Bern, Bern, Switzerland \and 
Graduate School for Cellular and Biomedical Sciences, Bern, Switzerland \and
Department of Diagnostic, Interventional, and Pediatric Radiology, Bern University Hospital, Bern, Switzerland
\and
Department of Radiology, Lucern Cantonal Hospital, Lucern, Switzerland
\and
Division of Pulmonology, Department of Medicine, Lausanne University Hospital (CHUV) and University of Lausanne, Lausanne, Switzerland
\begin{center}
\email{\{vasileios.dedousis, stavroula.mougiakakou\}@unibe.ch}
\end{center}}

\maketitle              
\let\thefootnote\relax
\footnotetext{\textsuperscript{\textdagger} Equal contribution.}


\begin{abstract}
Accurate segmentation of interstitial lung disease (ILD) patterns is essential for quantitative disease assessment and longitudinal monitoring. 
However, existing approaches remain limited by relying on dense annotations and producing static predictions that cannot be refined, motivating interactive approaches.
While promptable models show promise in interactive segmentation, their adaptation to ILDs remains largely unexplored.
To address this gap, we investigate prompt-guided foundation models for ILD refinement and present, to the best of our knowledge, the first adaptation of MedSAM2 for interactive 3D ILD segmentation on thoracic CT.
We investigate three fine-tuning strategies and multiple clinically motivated prompts: bounding-boxes (BBox), point, lasso, and scribble.
On a dataset spanning seven ILD patterns and healthy lung tissue, full model fine-tuning performed best, improving the average Dice score by 4.7 percentage points over MedSAM2.
While BBox prompts achieve the strongest performance, non-native MedSAM2 interactions such as lasso and scribble prompts also prove effective.
Finally, we present and evaluate a proof-of-concept end-to-end workflow in which MedSAM2 is initialized from an automatic segmentation prior and subsequently refined using radiologist prompts.
Model weights and plug-ins made available at: https://github.com/AIHNlab/ILD-SemiSegTool.



\end{abstract}

\section{Introduction}

Accurate thoracic computed tomography (CT) segmentation is vital for quantitative pulmonary imaging, supporting disease quantification, longitudinal monitoring, and computer-aided diagnosis \cite{maldonado2014automated,silva2018pulmonary,walsh2024deep}.
Deep learning methods, such as nnU-Net \cite{gao2026lung,isensee2021nnu,isensee2024nnu}, have advanced automated thoracic CT segmentation.
However, segmenting interstitial lung diseases (ILDs) remains considerably more challenging \cite{wasserthal2023totalsegmentator}.
ILD patterns such as ground-glass opacities (GGO), honeycombing (HC), and consolidation (Cons.) \cite{bankier2024fleischner} often exhibit fragmented spatial distributions, poorly defined boundaries, and substantial appearance variability \cite{travis2013official}. 
Unlike spatially coherent structures such as the lung lobes, which maintain a fixed number and relative location across patients despite locally ambiguous margins, ILD abnormalities may occur as a variable number of dispersed regions without fixed anatomical boundaries, complicating automated segmentation and motivating interactive refinement.
Methods have evolved from patch-based classification to dense semantic segmentation \cite{anthimopoulos2016lung,anthimopoulos2019semantic,christodoulidis2017multisource,fontanellaz2024computer} and to self-supervised learning \cite{choe2022content,dack2025unmasking}, but all require costly, inter-observer variable dense annotations or yield static predictions \cite{chassagnon2020deep,dack2023artificial,walsh2018deep,watadani2013interobserver}, motivating radiologist-led interactive refinement. 

Promptable models offer a promising framework for human-in-the-loop segmentation. 
The Segment Anything Model (SAM) established prompt-guided segmentation via points, bounding boxes (BBox), and masks \cite{kirillov2023segment}, while SAM2 introduced temporal memory to propagate masks across video frames \cite{ravi2024sam}.
This propagation mechanism is suited to volumetric CT, where inter-slice continuity enables sparse interactions to spread across a scan.
SAM-based architectures have already been applied to lung cancer lesion segmentation \cite{nguyen2026structsam,yi2026lungtumor}.
Although SAM2 has demonstrated strong performance under sparse volumetric interactions \cite{ulrich2024radioactive}, our preliminary experiments identified MedSAM2 \cite{ma2024medsam2} as the strongest baseline, motivating its use.
Existing medical adaptations largely target anatomical structures, leaving interactive segmentation of diffuse pathologies such as ILD patterns largely unexplored \cite{mazurowski2023segment,moraru2026semi}. 
Furthermore, most studies have been restricted to simple point and bounding-box prompts, with clinically motivated interactions such as scribbles (strokes over the target) and lassos (contours around the target) receiving limited attention, despite enabling rapid, intuitive region specification that mirrors freehand correction\cite{isensee2025nninteractive,ulrich2024radioactive}. 
Finally, it remains open whether automatic segmentations can initialize interactive correction of residual errors that scaling automatic models may not resolve, even though SAM-based frameworks support mask prompts \cite{khlaut2025radsam}.
As such, our contributions are threefold:

\begin{itemize}
    \item[---] Propose, to the best of our knowledge, the first adaptation of a SAM-based model (MedSAM2) for interactive ILD segmentation on thoracic CT.
    \item[---] Characterize how clinically motivated interactions support human-in-the-loop refinement of diffuse pulmonary abnormalities, including point, BBox, scribble, and lasso prompts (Fig.~\ref{fig:prompting}).
    \item[---] Present a proof-of-concept (PoC) workflow in which an automatic segmentation initializes radiologist-guided MedSAM2 refinement (Fig.~\ref{fig:pipeline}).

\end{itemize}

\section{Methods}

\subsection{Dataset}
The dataset \cite{bartholmai2006lung,christe2019computer} comprises 1 mm slice-thickness, hard-kernel-reconstructed CT scans from patients with diffuse fibrotic and obstructive parenchymal lung disease.
It includes 306 cases with ILD and lung masks, covering seven ILD patterns: GGO, HC, Cons., reticulation (Ret.), reticulation with GGO (Ret.+GGO), traction bronchiectasis (Bron.), emphysema (Emph.), and healthy tissue. 
Ground truth (GT) annotations came from two chest radiologists with 21 and 15 years of experience.
Ambiguities were resolved by consensus.
We split the dataset at the volume level as shown in \Cref{tab:dataset_split}. 
\begin{table}[t]
\centering
\caption{Case distribution of ILD patterns across dataset.}
\label{tab:dataset_split}
\setlength{\tabcolsep}{4pt}
\resizebox{0.7\linewidth}{!}{
\begin{tabular}{lcccccccc}
\toprule
Split & Healthy & GGO & Ret. & Cons. & HC & Ret.+GGO & Bron. & Emph. \\
\midrule
Train & 243 & 148 & 150 & 6 & 210 & 28 & 7  & 118 \\
Val   & 31  & 19  & 21  & 1 & 28  & 4  & 2  & 14  \\
Test  & 32  & 19  & 24  & 1 & 26  & 4  & 1  & 18  \\
Total & 306 & 186 & 195 & 8 & 264 & 36 & 10 & 150 \\
\bottomrule
\end{tabular}
}
\end{table}

\subsection{Segmentation}

\textbf{Automatic segmentation with nnU-Net}
nnU-Net \cite{isensee2021nnu,isensee2024nnu} is a self-configuring segmentation framework that automates the deep learning workflow and dynamically configures pre-processing pipelines, network topologies, and training schedules.
In our proposed framework, nnU-Net was used to generate an automated volumetric multi-class segmentation of the aforementioned ILD patterns for each CT volume, which we then use as the initial mask for the interactive refinement PoC workflow \cite{isensee2025nninteractive}.

\textbf{MedSAM2 adaptation.}
SAM2 \cite{ravi2024sam} is a promptable segmentation model that extends the SAM with a memory-based architecture for image/video segmentation. 
MedSAM2 \cite{ma2024medsam2} adapts SAM2 to the medical imaging domain by fine-tuning on diverse medical imaging datasets, initializing segmentation from prompts provided on a reference slice, and propagating predictions bidirectionally. 
We explored clinically relevant sparse interactions, such as point, BBox, scribble, and lasso prompts \cite{isensee2025nninteractive,ulrich2024radioactive,wong2024scribbleprompt}.
Although SAM2 natively supports points and BBox, MedSAM2 was pretrained exclusively with BBox prompts; we thus approximate scribble and lasso interactions as point series.
We split BBox into two types: a single-BBox (SBBox) enclosing the mask, and a multi-BBox (MBBox) where each disconnected region of a mask is assigned its own BBox.
For 3D propagation, we used the middle slice for initialization, following previous studies \cite{isensee2025nninteractive,ma2024medsam2}, and restricted selection to foreground slices where the target pattern was visible.
All prompts were derived from the GT mask as positive prompts marking regions to add, simulating radiologist interactions, assuming that selected regions correspond to correctly annotated target areas.

\begin{figure}[t]
    \centering

    \begin{subfigure}[t]{0.65\textwidth}
        \centering
        \includegraphics[width=\linewidth]{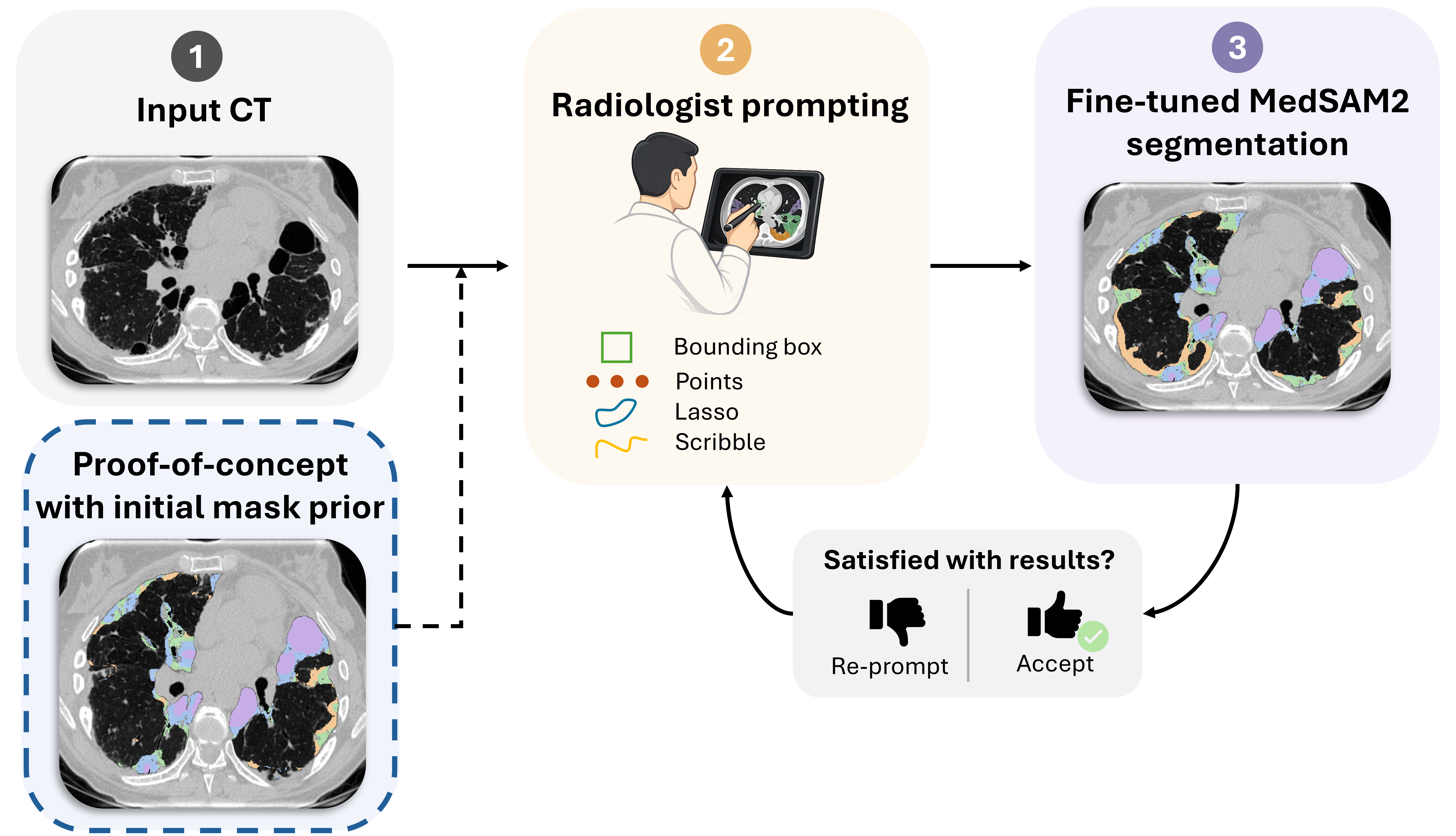}
        \caption{Interactive ILD segmentation workflow.}
        \label{fig:pipeline}
    \end{subfigure}
    \hfill
    \begin{subfigure}[t]{0.3\textwidth}
        \centering
        \includegraphics[width=\linewidth]{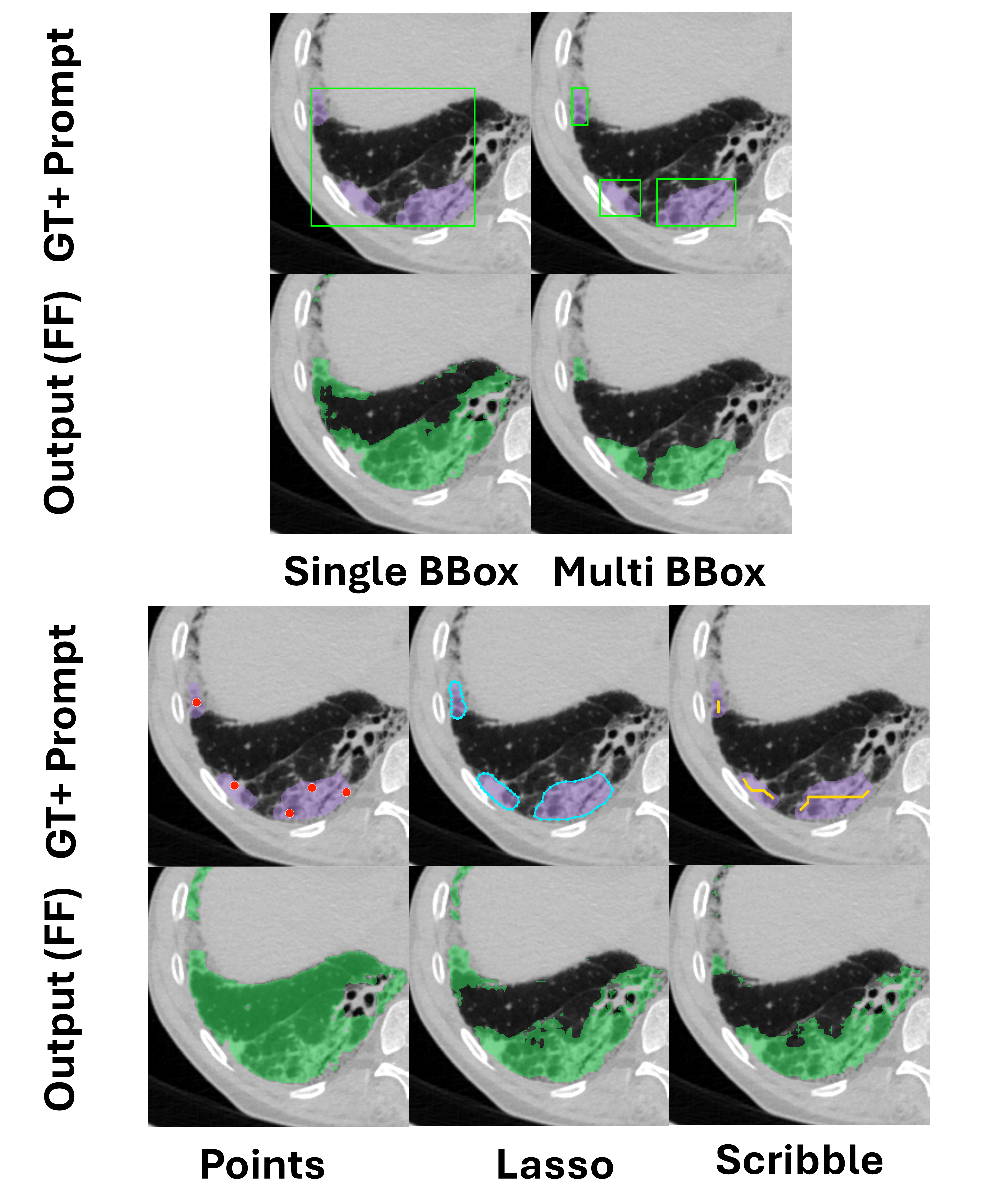}
        \caption{Prompt types.}
        \label{fig:prompting}
    \end{subfigure}

    \caption{Interactive segmentation framework. (a) Radiologist prompts drive FT MedSAM2 to segment CT images, optionally initialized from an automatic segmentation (PoC); prompts are simulated from ground-truth masks. (b) Prompts on a bronchiectasis slice, including single-BBox, multi-BBox, points, lasso, scribble, with GT on top row and FT MedSAM2 output on bottom row.}
    \label{fig:interactive_framework}
\end{figure}

\textbf{MedSAM2 with Mask Prior.}
For the PoC setting, nnU-Net trained for automatic ILD segmentation \cite{isensee2024nnu}, served as an initial mask prior (MP) for the downstream MedSAM2 interactive refinement workflow.
This was done to assess whether automatic predictions can serve as effective initialization priors for subsequent human-guided refinement. 
Starting from MP, a fine-tuned (FT) MedSAM2 performed refinement using the aforementioned prompt types. 
We opted to fully leverage prompt polarity under the SAM architecture: we used positive (P) for under- and negative (N) for over-segmented regions to indicate whether to include or exclude parts respectively.
Prompts were generated from discrepancies between nnU-Net predictions and GT masks. 
Negative BBox were excluded since they are not natively supported by SAM2.

\begin{table}[t]
\centering
\caption{Grid search for MedSAM2 fine-tuning. \textbf{Bold}, \underline{underline}, and \textit{italic} denote the optimal configuration for full finetuning, mask only, and mask and memory, respectively.}
\label{tab:medsam2_gridsearch}
\setlength{\tabcolsep}{6pt}
\resizebox{0.8\textwidth}{!}{
\begin{tabular}{l|l|l}
\hline
Learning rate & Weight decay & Loss weights \\
\hline
\textit{5e-5}, \textbf{1.7e-5}, 1.7e-4, \underline{5e-4} &
\textit{\underline{1e-4}}, \textbf{1e-1} &
$\{0.1, 1, 10\}$ orthogonal; \textbf{\underline{(1,1,1)}}, \textit{(10,10,10)} \\
\hline
\end{tabular}
}
\end{table}

\subsection{Experimental Setup and Evaluation}
All experiments ran on NVIDIA RTX 4090.
For semi-automatic segmentation, models were initialized from the MedSAM2\_2411 checkpoint and FT under three strategies: mask decoder only (MO), mask decoder and memory modules (MM), and full fine-tuning (FF). 
Training used eight frames \cite{ma2024medsam2}, sampling one interaction type per epoch and rotating modalities across epochs to expose the model to every prompt type. 
To limit hyperparameter search, we equally weighted the Dice and focal losses and used identical learning rates across model components.

Hyperparameter optimization used the gridsearch in \Cref{tab:medsam2_gridsearch}, performed only on the standalone MedSAM2 setting.
Models were trained with batch size four, using a cosine learning rate schedule with 10\% linear warmup. 
The objectness classification loss was fixed at unit weight.
The best configuration for each fine-tuning strategy was chosen by validation loss, reached at epochs 47, 52, and 52, respectively, for MO, MM, and FF.
For the best-performing FF model, we tested sensitivity to the 3D initialization slice, offset from the mid-slice by ±5–50\%, averaged across prompt types (\Cref{fig:init_average}).

As FF consistently performed best, all subsequent PoC experiments with the dense mask prior (MP) used FF with hyperparameters from the standalone grid search (best epoch 52).
For the PoC, we trained nnU-Net v2 [14] on our ILD dataset using its self-configuring framework, reaching the best epoch at 991.
Its predictions were then used to further fine-tune MedSAM2 with the MP.
To assess prompt polarity, we evaluated P-only, N-only, and combined P+N refinement.

We evaluated segmentation performance using the Dice similarity coefficient (DSC) and normalized surface distance (NSD, 2\,mm tolerance) to assess boundary agreement \cite{ma2024medsam2,maier2024metrics}.
Following hierarchy-aware principles \cite{maier2024metrics}, metrics were averaged across classes within each volume then across all volumes.
We compare against vanilla SAM2 (SAM2.1-Tiny) \cite{ravi2024sam}, MedSAM2 (MedSAM2\_2411) \cite{ma2024medsam2}, and nnInteractive (v1.0) \cite{isensee2025nninteractive}, an interactive medical image segmentation framework built upon the nnU-Net architecture.
The latter was evaluated in two settings: without an initial mask, for direct comparison with FT MedSAM2, and with MP derived from nnU-Net predictions, for comparison with PoC.

\section{Results and Discussion}

\subsection{MedSAM2 adaptation}

\textbf{Module adaptation.}
\Cref{tab:3d_results_mid} summarizes 3D performance for vanilla SAM, MedSAM2, nnInteractive, and the three fine-tuning strategies across prompt types. 
Despite previous reports favoring SAM2, MedSAM2 outperformed SAM2 in most cases, motivating its selection for fine-tuning. 
FF achieved the highest DSC (0.380) and NSD among standalone models across all prompt types.
The NSD gains indicate sharper boundary delineation rather than volumetric overlap alone. 
This is a key distinction given the poorly defined margins of ILD patterns.
MM underperformed MO on most interactions, indicating that fine-tuning the memory module alone provides no benefit and that the image encoder, adapted only in FF, is the critical factor in capturing the heterogeneous appearance and distribution of ILD patterns.
These results established FF as the best adaptation strategy and motivated its use for subsequent experiments.

\textbf{Best prompt strategy analysis.} Among the evaluated interactions, MBBox produced the highest performance, consistent with ILD's multifocal nature, which a single enclosing box often fails to capture. 
Lasso and scribble, though absent from MedSAM2 pretraining, achieved performance comparable to native point prompts, suggesting that the model accommodates richer, clinically motivated interactions without explicit architectural changes.
Unlike the MedSAM2 and nnInteractive baselines, our FF model maintained stable and statistically superior performance across all five prompt types (\Cref{tab:3d_results_mid}).
This is reflected qualitatively in \Cref{fig:prompting}, where FT MedSAM2 outputs closely follow GT boundaries.
nnInteractive collapsed on lasso (0.041) and point prompts (0.195), while performing comparatively better on MBBox (0.398) and scribble (0.376). 
The latter is partly driven by Bronc. (\Cref{tab:per_class_dsc_baselines}), interpreted cautiously, given its single test case. 
Our FF model maintained consistently higher DSC and NSD across all five interaction types, reflecting greater robustness to prompt variability.

\begin{table}[t]
\centering
\caption{Performance across prompt types (3D, \textit{mid-slice init}) comparing baselines and standalone fine-tuning strategies: MO, MM, and FF. Best in \textbf{bold}, second best \uline{underlined}. \textcolor{blue}{$\star$}/\textcolor{blue}{$\triangledown$}: significantly better/worse than MedSAM2; \textcolor{orange!90!black}{$\star$}/\textcolor{orange!90!black}{$\triangledown$}: significantly better/worse than nnInteractive (paired Wilcoxon signed-rank test, case-averaged, Holm--Bonferroni corrected, $p<0.05$). Unmarked cells: no statistical significance.}
\label{tab:3d_results_mid}
\setlength{\tabcolsep}{4pt}
\resizebox{\textwidth}{!}{
\begin{tabular}{lcccccccccc}
\toprule
& \multicolumn{2}{c}{SBBox}
& \multicolumn{2}{c}{MBBox}
& \multicolumn{2}{c}{Points}
& \multicolumn{2}{c}{Lasso}
& \multicolumn{2}{c}{Scribble} \\
\cmidrule(lr){2-3}
\cmidrule(lr){4-5}
\cmidrule(lr){6-7}
\cmidrule(lr){8-9}
\cmidrule(lr){10-11}
Model & DSC & NSD & DSC & NSD & DSC & NSD & DSC & NSD & DSC & NSD \\
\midrule
SAM 2.1       & 0.255 & 0.206 & 0.340 & 0.319 & 0.183 & 0.094 & 0.213 & 0.133 & 0.280 & 0.226 \\
MedSAM2       & 0.347 & 0.323 & 0.337 & 0.314 & 0.333 & 0.298 & 0.359 & 0.322 & 0.318 & 0.280 \\
nnInteractive & 0.333 & 0.273 & 0.398 & 0.313 & 0.195 & 0.181 & 0.041 & 0.074 & 0.376 & 0.325 \\
\midrule
MO (Ours) & \uline{0.383}\bstar\ystar & \uline{0.349}\ystar & \uline{0.418}\bstar\ystar & 0.365\bstar\ystar & \uline{0.355}\bstar\ystar & \uline{0.316}\ystar & 0.399\bstar\ystar & \uline{0.348}\bstar\ystar & 0.383\bstar & 0.333 \\
MM (Ours) & 0.373\ystar & 0.327\ystar & \uline{0.418}\bstar\ystar & \uline{0.374}\bstar\ystar & 0.349\ystar & 0.301\ystar & \uline{0.400}\bstar\ystar & 0.344\ystar & \uline{0.396}\bstar & \uline{0.346}\bstar\ystar \\
FF (Ours) & \textbf{0.414}\bstar\ystar & \textbf{0.381}\bstar\ystar & \textbf{0.450}\bstar\ystar & \textbf{0.413}\bstar\ystar & \textbf{0.421}\bstar\ystar & \textbf{0.381}\bstar\ystar & \textbf{0.426}\bstar\ystar & \textbf{0.379}\bstar\ystar & \textbf{0.425}\bstar\ystar & \textbf{0.380}\bstar\ystar \\
\bottomrule
\end{tabular}}
\end{table}

\begin{figure}[H]
    \centering

    \begin{subfigure}[t]{0.55\textwidth}
        \centering
        \includegraphics[width=0.85\linewidth]{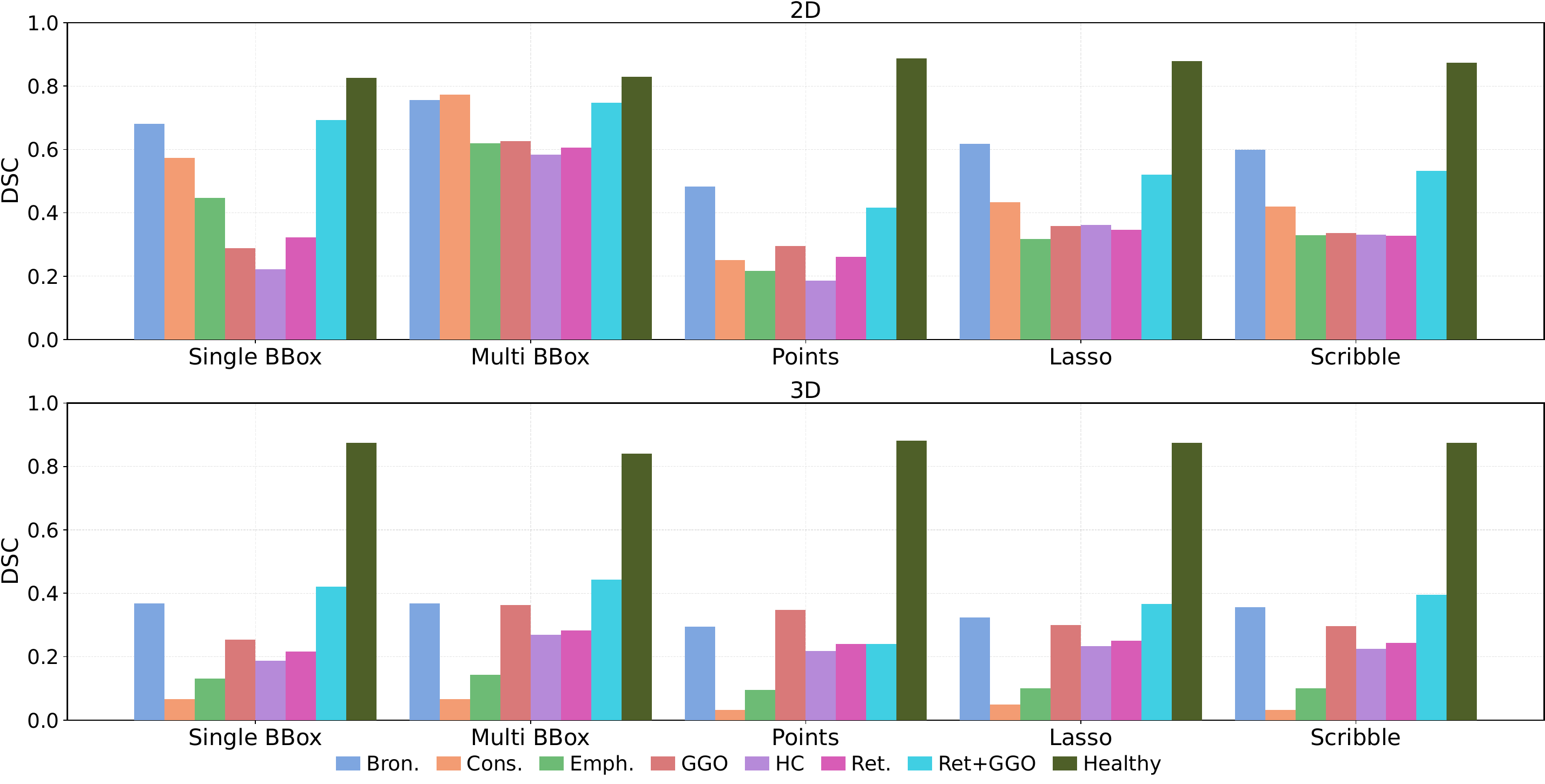}
        \caption{FF MedSAM2 per-pattern DSC score across prompt types in 2D and 3D.}
        \label{fig:barplot}
    \end{subfigure}
    \hfill
    \begin{subfigure}[t]{0.4\textwidth}
        \centering
        \includegraphics[width=0.85\linewidth]{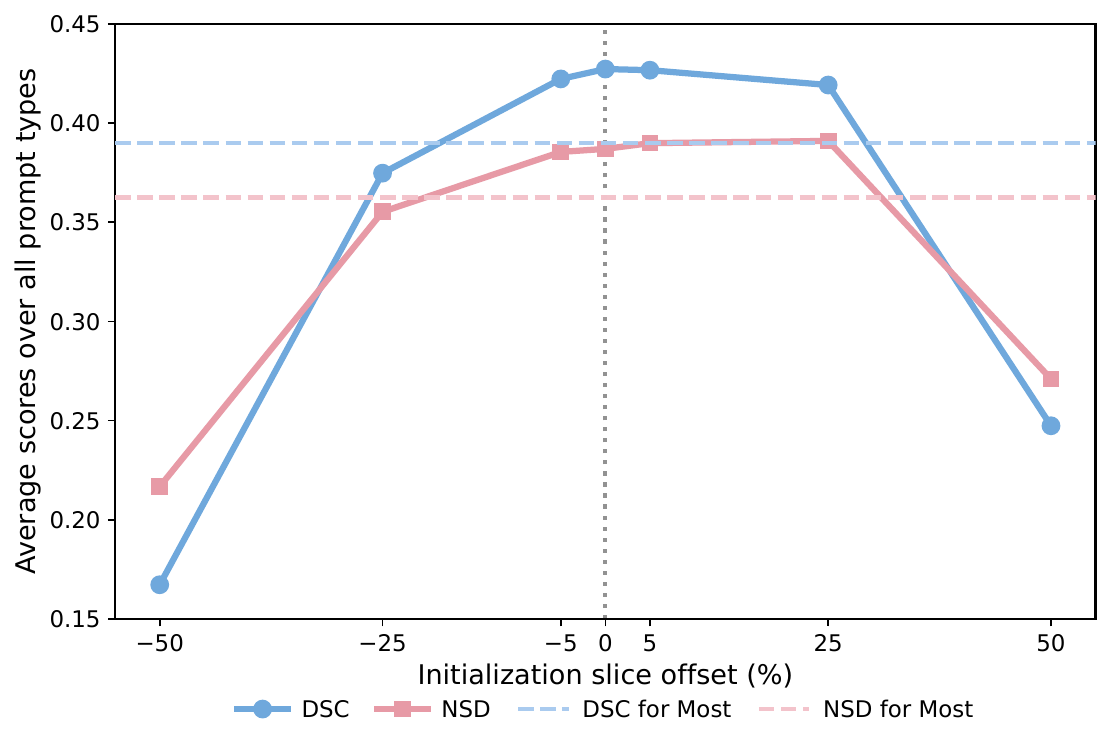}
        \caption{FF sensitivity to 3D initialization slice averaged over prompts.}
        \label{fig:init_average}
    \end{subfigure}

   \caption{Evaluation of FF MedSAM2. (a) Per-pattern DSC across prompt types in 2D and 3D. (b) Sensitivity of 3D performance to initialization slice: average DSC and NSD versus offset from the middle ground-truth slice (0\%); dashed lines indicate initialization from the ground-truth slice with the most foreground.}
    \label{fig:performance_results}
\end{figure}

\begin{table}[t]
\centering
\caption{Per-pattern DSC of baseline models nnInteractive and MedSAM2.}
\label{tab:per_class_dsc_baselines}
\setlength{\tabcolsep}{4pt}
\resizebox{0.8\textwidth}{!}{
\begin{tabular}{llccccccccc}
\toprule
Model & Prompt & Bron. & Cons. & Emph. & GGO & HC & Ret. & Ret.+GGO & Healthy & Case Avg. \\
\midrule
\multirow{5}{*}{MedSAM2}
& SBBox    & 0.128 & 0.610 & 0.177 & 0.131 & 0.068 & 0.149 & 0.352 & 0.832 & 0.347 \\
& MBBox    & 0.128 & 0.610 & 0.157 & 0.140 & 0.075 & 0.174 & 0.365 & 0.777 & 0.337 \\
& Points   & 0.139 & 0.043 & 0.148 & 0.155 & 0.059 & 0.143 & 0.232 & 0.831 & 0.333 \\
& Lasso    & 0.139 & 0.043 & 0.184 & 0.192 & 0.099 & 0.202 & 0.236 & 0.809 & 0.359 \\
& Scribble & 0.133 & 0.043 & 0.173 & 0.155 & 0.087 & 0.158 & 0.221 & 0.733 & 0.318 \\
\midrule
\multirow{5}{*}{nnInteractive}
& SBBox    & 0.480 & 0.553 & 0.249 & 0.115 & 0.064 & 0.112 & 0.218 & 0.790 & 0.333 \\
& MBBox    & 0.482 & 0.553 & 0.336 & 0.168 & 0.095 & 0.172 & 0.409 & 0.850 & 0.398 \\
& Points   & 0.400 & 0.432 & 0.158 & 0.106 & 0.064 & 0.136 & 0.254 & 0.336 & 0.195 \\
& Lasso    & 0.022 & 0.025 & 0.171 & 0.036 & 0.042 & 0.051 & 0.029 & 0.007 & 0.041 \\
& Scribble & 0.401 & 0.440 & 0.253 & 0.169 & 0.090 & 0.153 & 0.271 & 0.853 & 0.376 \\
\midrule
\end{tabular}
}
\end{table}

\textbf{Prompting strategies}. \Cref{fig:barplot} compares FF under 2D interactions, where prompts are provided on every slice, and 3D interactions, where prompts are limited to one slice and propagated bidirectionally. 
Across most patterns, 2D higher DSC was achieved over 3D, indicating a performance cost associated with reduced interaction burden.
MBBox remained the most robust prompt in both settings, whereas points showed the greatest 2D-to-3D degradation across several patterns.
FF performance under different prompts (\Cref{fig:barplot}) reveals that MBBox dominance does not hold uniformly across patterns.
For fragmented, diffuse patterns such as GGO, HC, Emph., and Ret., MBBox outperforms all others, nearly doubling DSC relative to SBBox and point-based interactions, consistent with their irregular, multifocal distribution.
For healthy tissue, point, lasso, and scribble prompts achieve comparable or slightly higher DSC than either BBox variant, as all three reduce to point-based seeding under our approximation, sufficient for this large, continuous structure. 
These results indicate that optimal prompt type is pattern-dependent: MBBox was best for fragmented multifocal patterns, whereas point-based was best for large, contiguous structures.
FF performance peaked near the mid-slice and degraded toward the volume extremities (\Cref{fig:init_average}), with initialization from the GT slice of greatest foreground showing no substantial gains, supporting the mid-slice as the choice.
Clinically, this requires identifying a pathology-containing slice before interaction; initializing on less representative slices degrades performance (Fig.~\ref{fig:init_average}) motivating future work on automatic slice selection or localization-free workflows.

\begin{table}[t]
\centering
\caption{Per-pattern DSC across prompt types for PoC pipeline. Best per column in \textbf{bold}, second best \uline{underlined}. Significance reported on Case Avg.\ only (paired Wilcoxon, 32 volumes, Holm--Bonferroni corrected): \textcolor{blue}{$\star$}/\textcolor{blue}{$\triangledown$} significantly better/worse than nnU-Net; \textcolor{orange!90!black}{$\star$}/\textcolor{orange!90!black}{$\triangledown$} significantly better/worse than nnInteractive+MP. Unmarked: no statistical significance. Case Avg.\ averaged across prompts: 0.448 (FF+MP P+N), 0.434 (FF+MP P), 0.501 (FF+MP N), 0.303 (nnInteractive+MP).}
\label{tab:per_class_dsc_mask}
\setlength{\tabcolsep}{4pt}
\resizebox{0.9\textwidth}{!}{
\begin{tabular}{llccccccccc}
\toprule
Model & Prompt & Bron. & Cons. & Emph. & GGO & HC & Ret. & Ret.+GGO & Healthy & Case Avg. \\
\midrule
nnU-Net & -- & 0.039 & 0.006 & 0.346 & \textbf{0.526} & \textbf{0.464} & \uline{0.397} & 0.276 & \textbf{0.922} & 0.504 \\
\midrule
\multirow{5}{*}{nnInteractive + MP}
& SBBox    & 0.073 & 0.006 & 0.283 & 0.165 & 0.093 & 0.124 & 0.134 & \uline{0.869} & 0.291 \\
& MBBox    & 0.085 & 0.006 & 0.346 & 0.178 & 0.166 & 0.197 & 0.240 & 0.700 & 0.304 \\
& Points   & 0.074 & 0.006 & 0.297 & 0.191 & 0.113 & 0.167 & 0.200 & 0.805 & 0.300 \\
& Lasso    & 0.055 & 0.006 & 0.306 & 0.204 & 0.165 & 0.191 & 0.239 & 0.802 & 0.323 \\
& Scribble & 0.084 & 0.006 & 0.323 & 0.155 & 0.125 & 0.152 & 0.246 & 0.788 & 0.299 \\
\midrule
\multirow{5}{*}{FF+MP (P+N) (Ours)}
& SBBox    & 0.304 & 0.084 & 0.347 & 0.363 & 0.183 & 0.340 & \textbf{0.419} & 0.865 & 0.460\ystar \\
& MBBox    & \textbf{0.345} & 0.084 & 0.338 & 0.445 & 0.376 & \textbf{0.426} & \uline{0.413} & 0.794 & \textbf{0.508}\ystar \\
& Points   & 0.181 & 0.100 & 0.315 & 0.356 & 0.190 & 0.335 & 0.305 & 0.825 & 0.439\ystar \\
& Lasso    & 0.177 & 0.208 & 0.179 & 0.249 & 0.166 & 0.248 & 0.388 & 0.847 & 0.413 \\
& Scribble & 0.177 & \textbf{0.217} & 0.234 & 0.227 & 0.195 & 0.248 & 0.356 & 0.847 & 0.419\ystar \\
\midrule
\multirow{5}{*}{FF+MP (P) (Ours)}
& SBBox    & 0.304 & 0.084 & 0.347 & 0.363 & 0.183 & 0.340 & \textbf{0.419} & 0.865 & 0.460\ystar \\
& MBBox    & \textbf{0.345} & 0.084 & 0.338 & 0.445 & 0.376 & \textbf{0.426} & \uline{0.413} & 0.794 & \textbf{0.508}\ystar \\
& Points   & 0.181 & 0.166 & 0.167 & 0.299 & 0.127 & 0.278 & 0.242 & 0.868 & 0.405\ystar \\
& Lasso    & 0.177 & 0.181 & 0.150 & 0.237 & 0.153 & 0.234 & 0.243 & 0.842 & 0.397 \\
& Scribble & 0.177 & \uline{0.214} & 0.168 & 0.237 & 0.153 & 0.228 & 0.298 & 0.843 & 0.400\ystar \\
\midrule
\multirow{3}{*}{FF+MP (N) (Ours)}
& Points   & \uline{0.315} & 0.027 & \textbf{0.417} & 0.426 & \uline{0.378} & 0.353 & 0.404 & 0.838 & 0.505\ystar \\
& Lasso    & \uline{0.315} & 0.027 & \uline{0.389} & 0.465 & 0.271 & 0.353 & 0.405 & 0.852 & 0.491\ystar \\
& Scribble & \uline{0.315} & 0.025 & 0.364 & \uline{0.472} & 0.325 & 0.349 & 0.402 & 0.851 & \uline{0.506}\ystar \\
\bottomrule

\end{tabular}
}
\end{table}

\subsection{MedSAM2 with Mask Prior}

\textbf{Prompt ablation.}
Introducing the automatic MP substantially improved performance over the non-MP setting (\Cref{tab:3d_results_mid}) across all interaction types (\Cref{tab:per_class_dsc_mask}). 
Averaged across prompts, FF+MP (P+N) achieved DSC of 0.448, exceeding nnInteractive coupled with nnU-Net MP (DSC: 0.303) with statistical significance.
FF+MP (N) scored highest among point-based interactions, while FF+MP (P) consistently underperformed. 

\textbf{Per-pattern and prompt polarity analysis.}
The best coupled configuration was FF+MP (P+N) with MBBox prompts (0.508), with no statistical significance over baselines, while FF+MP (N) with scribble (0.506) was the strongest point-based alternative (\Cref{tab:per_class_dsc_mask}). 
Despite this, nnU-Net remained superior for GGO, HC, and healthy tissue, where its per-voxel predictions closely matched the GT, leaving little systematic error for the prompts to correct.
The coupled pipeline yielded clearer gains for Ret. and Ret.+GGO, with finer, multifocal patterns prone to under-segmentation, where clician corrections can directly recover missing structures.
These effects depended on polarity: negative-only prompts preferentially improved Emph., GGO, and HC, consistent with these patterns being prone to over-segmentation by nnU-Net, while positive-only prompts performed best for Ret. and Ret.+GGO, suggesting under-segmentation. 
Combining both polarities did not consistently outperform single-polarity refinement, though it achieved the highest DSC in Bronc., interpreted cautiously given its under-representation.
Overall, the coupled pipeline improves fragmented, erroneously segmented patterns while trading performance on patterns nnU-Net already segments well, demonstrating complementary rather than uniformly superior performance consistent with PoC characterization.

\subsection{Limitations and Future Work}

While this work establishes the first adaptation of MedSAM2 for interactive ILD segmentation and demonstrates the potential of prompt-guided refinement, several limitations warrant consideration.
The dataset exhibits substantial class imbalance, particularly for Cons. and Bron., rendering per-pattern results unreliable.
This modest DSC reflects task difficulty rather than a method-specific limitation, as comparable baselines show similar or lower performance (\Cref{tab:3d_results_mid}); clinical utility as an interactive aid would still require prospective validation. Validation-based model selection and stratified splitting make overfitting unlikely, though training-time class balancing was not applied and remains worth exploring.
Interactions were simulated based on ground-truth annotations and therefore do not fully capture the variability and imprecision of real user input.
Moreover, the PoC experiments were initialized from a single automatic segmentation model and a single round of interactive corrections, leaving iterative multi-round refinement and generalizability across initialization strategies unexplored.
Finally, propagation errors may accumulate for irregularly distributed patterns despite 3D interactions reducing annotation burden.
Future work should therefore evaluate the proposed framework prospectively with radiologists, investigate uncertainty-aware and iterative correction strategies, assess robustness across external datasets and alternative automatic initialization models and better exploit non-native interactions, such as lasso and scribble prompts, while improving memory propagation for highly fragmented ILD patterns.

\section{Conclusion}
To the best of our knowledge, we present the first adaptation of MedSAM2 for interactive 3D CT segmentation of ILD patterns.
Full model fine-tuning consistently achieved the strongest performance, suggesting full adaptation best addresses the domain shift posed by diffuse pulmonary abnormalities. 
While 2D slice-level interactions outperformed volumetric 3D propagation, the latter retained reasonable performance at substantially lower interaction burden, highlighting both the potential and limitations of memory-based propagation for irregularly distributed ILD patterns. 
We further demonstrated, as a PoC, that dense prompting for SAM-based models, such as automatic segmentation, can serve as promising initialization priors for subsequent MedSAM2 prompt-guided refinement. 
Together, these findings establish the feasibility of interactive foundation models for ILD segmentation suggesting their greatest value may lie not in replacing automatic algorithms but in enabling clinically meaningful human-in-the-loop refinement for quantitative thoracic imaging.

\begin{credits}
\textbf{\ackname} 
This work was partly supported by the Swiss National Science Foundation through the project PRISM-fILD (10006062). All computations were performed on UBELIX (https://www.id.unibe.ch/hpc), the HPC cluster at the University of Bern.
We are grateful to Md. Mahadi Hassan Munna, Research Software Engineer, for his contribution in developing the custom OHIF plugin that integrated our medical image annotation algorithms into the viewing platform.

\textbf{\discintname}
The authors have no competing interest to declare.
\end{credits}
%
%
%
\bibliographystyle{splncs04}
\bibliography{ref}
%




\end{document}